# APPLICATION OF THRESHOLD TECHNIQUES FOR READABILITY IMPROVEMENT OF JAWI HISTORICAL MANUSCRIPT IMAGES

Hafizan Mat Som[1], Jasni Mohamad Zain[2] and Amzari Jihadi Ghazali[3]

[1]IKIP International College Taman Gelora, 25250 Kuantan, Pahang, Malaysia
[2]Faculty Of Computer Systems and Software Engineering, University Malaysia Pahang, Lebuhraya Tun Razak, 26300, Gambang, Kuantan, Pahang, Malaysia
[3]Kolej Profesional Mara Indera Mahkota, Jln Sg Lembing, 25200 Kuantan, Pahang, Malaysia
email : [1]hafizan@ikipskills.edu.my, [2]jasni@ump.edu.my, [3]ajihadi@mail.mara.go.my

**ABSTRACT:**

*Abstract: Problem statement: Historical documents such as old books and manuscripts have a high esthetic value and highly appreciated. Unfortunately, there are some documents cannot be read due to quality problems like faded paper, ink expand, uneven color tone, torn paper and other elements disruption such as the existence of small spots. The study aims to produce a copy of manuscript that shows clear wordings so they can easily be read and the copy can also be displayed for visitors. Approach: 16 samples of Jawi historical manuscript with different quality problems were obtained from The Royal Museum of Pahang, Malaysia. We applied three binarization techniques; Otsu's method represents global threshold technique; Sauvola and Niblack method which are categorized as local threshold techniques. There are also a pre processing step involving histogram equalization process, morphology functions and filtering technique. Finally, we compare the binarized images with the original manuscript to be visually inspected by the museum's curator. The unclear features were marked and analyzed. Results: Most of the examined images show that with optimal parameters and effective pre processing technique, local thresholding methods are work well compare with the other one. Even the global thresholding method give less cost in computational time, the results were not yet satisfied. Niblack's and Sauvola's techniques seem to be the suitable approaches for these types of images. Most of binarized images with these two methods show improvement for readability and character recognition. For this research, even the differences of image result were hard to be distinguished by human capabilities, after comparing the time cost and overall achievement rate of recognized symbols, Niblack's method is performing better than Sauvola's. Conclusion: There is no single algorithm that works well for all types of images but some work well than others for particular types of images suggesting that improved performance can be obtained by automatic selection or combination of appropriate algorithms for the type of document image under investigation. We could improve the post processing step by adding edge detection techniques and further enhanced by an innovative image refinement technique and a formulation of a class proper method.*

## 1. INTRODUCTION

Historical documents such as old books and manuscripts have a high esthetic value and highly appreciated[1]. These documents must always be kept and conserve in good condition by organization like museum and national archive because it contains information and messages that become the evidence of human existence and a living community in the past.

Unfortunately, there are some documents cannot be read due to quality problems like faded paper, ink expand, uneven color tone, torn paper and other elements disruption such as the existence of small spots and others as shown in Figure 1.

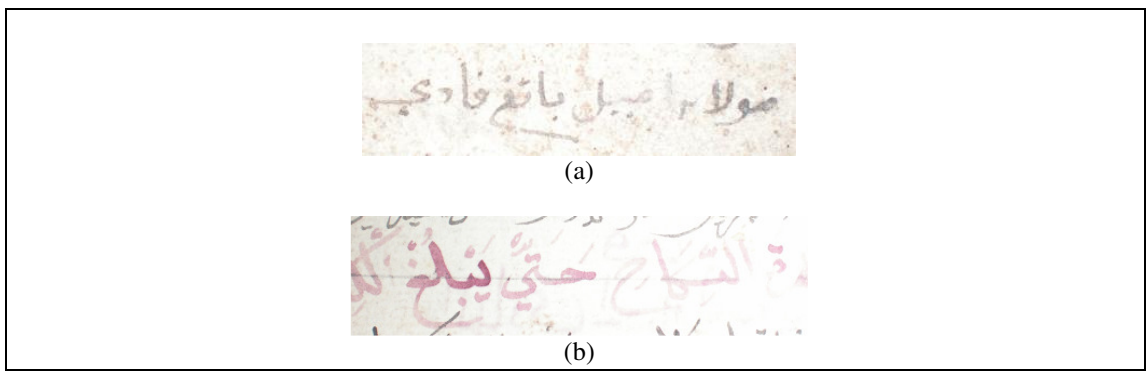

(a)

(b)

*Figure 1 : (a) Sample of Image with existence of small spots (b) Sample of Image with uneven ink tones*

It is very important to preserve these kind of artefact as it were become references, research resources and learning materials for grown civilization[15]. Some original manuscripts were prohibited to be displayed to public due to their condition or confidential issues. But, to show and expose these materials in other form is become first priority for the museum, so people can learn about their predecessors.

Therefore, this paper focuses to convert the degraded document image to a new form that is binary image, which means a black and white image. The study aims to produce a copy of document that shows clear wordings so they can implement further studies and the copy also can be display for visitors. Image binarization is a technique that uses a threshold value to convert from colour to binary image. It is able to separate the foreground and background so the segmented wordings can be seen clearly. In this paper, we applied three types of techniques that is Otsu's method representing Global threshold also Sauvola's and Niblack method which are categorized as Local threshold. There are also a pre processing step involving histogram equalization process, morphology functions and filtering technique.

Finally, we compare the binarized images with the original manuscript to be visually inspected by the expert museum's curator. The unclear features were marked and analyzed to be decided which method is the best.

## 2. RELATED WORK

Histogram Equalization

Histogram equalization provide a sophisticated method for modifying the dynamic range and contrast of an image by altering that image such that its intensity histogram has a desired shape[2]. Unlike contrast stretching, histogram modeling operators may employ non-linear and non-monotonic transfer functions to map between pixel intensity values in the input and output images. Histogram equalization employs a monotonic, non-linear mapping which re -assigns the intensity values of pixels in the input image such that the output image contains a uniform distribution of intensities (i.e. a flat histogram)[2].Figure 2 shows example of histogram equalization process.

## Otsu's Method

Global thresholding uses only one threshold value, which is estimated based on statistics or heuristics on global image attributes, to classify image pixels into foreground or background. The major drawback of global thresholding techniques is that it cannot differentiate those pixels which share the same gray level but do not belong to the same group[3]. Otsu's method [4] is one of the best global thresholding methods. It works well with dearly scanned images, but it performs unsatisfactorily for those poor quality images that have low contrast and non-uniform illumination.

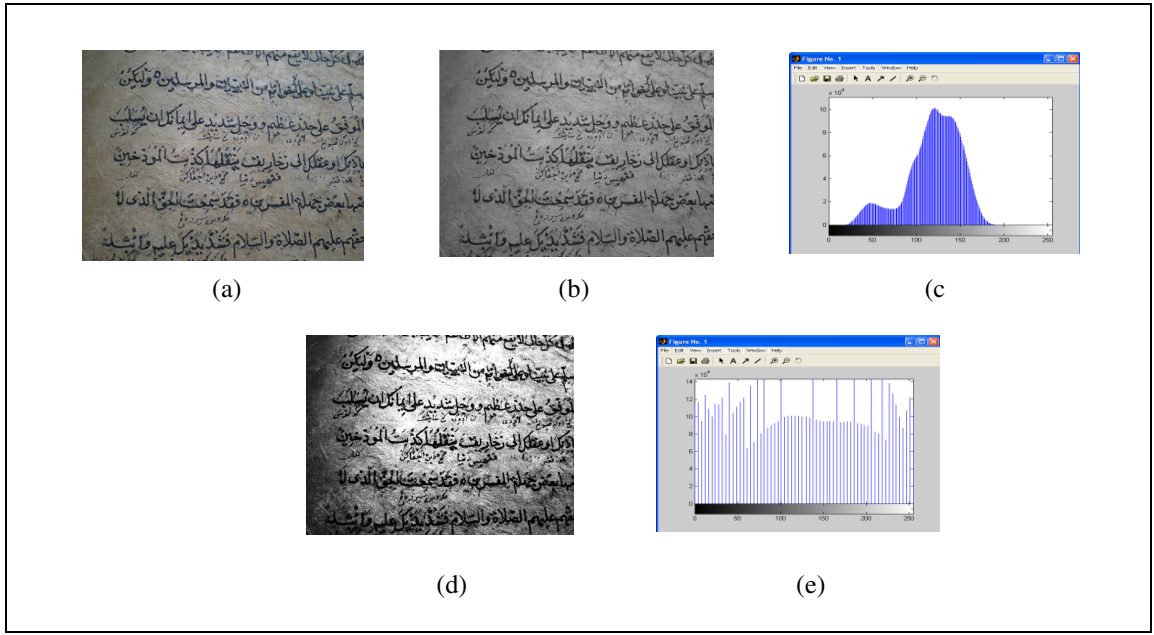

(a)  (b)  (c)

(d)  (e)

*Figure 2 : (a) Original RGB Image (b) Grayscale Image (c) Histogram of Grayscale Image (d) Equalized Image (e) Histogram of Equalized Image*

Otsu's thresholding method involves iterating through all the possible threshold values and calculating a measure of spread for the pixel levels each side of the threshold, i.e. the pixels that either fall in foreground or background. The aim is to find the threshold value where the sum of foreground and background spreads is at its minimum. The calculations for finding the foreground and background variances (the measure of spread) for a single threshold are as follows:-

1.  Find the threshold that minimizes the weighted within-class variance.
2.  This turns out to be the same as maximizing the between-class variance.
3.  Operates directly on the gray level histogram
4.  The weighted within-class variance is:

$$\sigma_w^2(t) = q_1(t)\sigma_1^2(t) + q_2(t)\sigma_2^2(t) \qquad (1)$$

Where the class probabilities are estimated as:

$$q_1(t) = \sum_{i=1}^{t} P(i) \qquad q_2(t) = \sum_{i=t+1}^{I} P(i)$$

(2)

And the class means are given by:

$$\mu_1(t) = \sum_{i=1}^{t} \frac{iP(i)}{q_1(t)} \qquad \mu_2(t) = \sum_{i=t+1}^{I} \frac{iP(i)}{q_2(t)}$$

(3)

Finally, the individual class variances are:

$$\sigma_1^2(t) = \sum_{i=1}^{t} [i - \mu_1(t)]^2 \frac{P(i)}{q_1(t)}$$

$$\sigma_2^2(t) = \sum_{i=t+1}^{I} [i - \mu_2(t)]^2 \frac{P(i)}{q_2(t)}$$

(4)

5.    Run through the full range of t values [1,256] and pick the value that minimizes. However, it is difficult to achieve consistent quality with a fixed threshold while processing a batch of archive images, because both foreground and background colours vary significantly between images.

Niblack's Algorithm

In local thresholding, the threshold values are spatially varied and determined based on the local content of the target image. In comparison with global techniques, local tresholding techniques have better performance against noise and error especially when dealing with information near texts or objects. According to Trier's survey[5], Yanowitz Bruckstein's method[6] and Niblacks method[7] are two of the best performing local thresholding methods. Yanowitz-Bruckstein's method is extraordinary complicated and thus requires very Iarge computational power. This makes it infeasible and too expensive for real system implementations. On the other hand, Niblacks method is simple and effective. As a result, we decided to focus on Niblack's method. Niblack's algorithm [7] is a local thresholding method based on the calculation of the local mean and of local standard deviation. The threshold is decided by the formula:

$$T(x, y) = m(x, y) + k \bullet s(x, y),$$

(5)

where m(x, y) and s(x, y) are the average of a local area and standard deviation values, respectively. The size of the neighborhood should be small enough to preserve local details, but at the same time large enough to suppress noise. The value of k is used to adjust how much of the total print object boundary is taken as a part of the given object. Zhang and Tan [8] proposed an improved version of Niblack's algorithm:

$$T(x, y) = m(x, y) \bullet \left[ 1 + k \bullet \left( 1 - \frac{s(x, y)}{R} \right) \right]$$

(6)

where k and R are empirical constants. The improved Niblack method[8] uses parameters k and R to reduce its sensitivity to noise.

Sauvola's Algorithm

Sauvola's algorithm [9] claims to improve Niblack's method by computing the threshold using the dynamic range of image gray-value standard deviation, R:

$$T_{Sauvola} = m * (1 - k * (1 - \frac{s}{R}))$$ (7)

where k is set to 0.5 and R to 128. This method outperforms Niblack's algorithm in images where the text pixels have near 0 gray value and the background pixels have near 255 gray-values. However, in images where the gray values of text and non-text pixels are close to each other, the results degrade significantly.

The Figure 3 below shows the methodology for this research. There are five phases involved; Image acquisition, data definition, image pre-processing, image binarization and evaluation. The images of manuscript are varied in terms of its sizes, color tone condition and fonts for the testing process (during binarization).

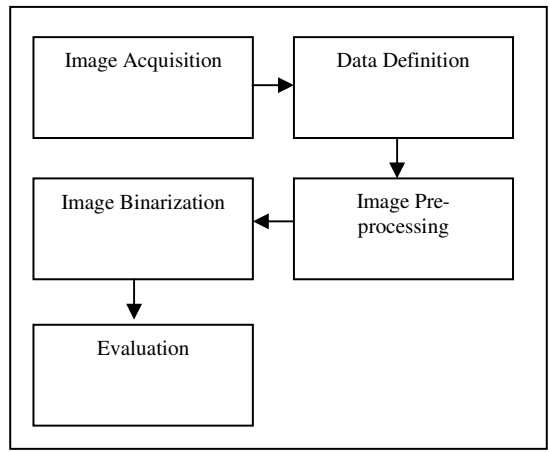

*Figure 3: The methodology of research*

The images are acquired by a digital camera with high resolution ratio, stored in computer and converted to compressed file formats with low storage requirements. The digital camera used is a 7.1 megapixel Olympus FE-230 with a 2.5 inch LCD. The maximum size of an image is 3072x2304 pixels[10]. JPEG files were used due to lower size and computation time, especially for old books reproduction, since RAW or TIFF files are memory and CPU time consuming. "MATLAB IP Toolbox" [2004] supports these entire formats. We worked with a sufficient number of document images and applied various filters, extended too many types, sizes, windows, etc, in order to explore denoising procedures.

Data that comes from the form of images of manuscript need to be identified and analyzed first before tested. The information about color tone condition and degraded wordings need to be gathered since our problem is different types of quality distortion occur in the manuscripts which lead to problem in recognizing characters by human view. Therefore, by having this information, all data to be tested will obtain full attention in terms of pre processing techniques applied. The preparation of the image conducted before filtering and binarization consists of conversion from RAW to TIFF / JPEG and cropping.

Three processes are involved in pre-processing which are converting image to grayscale, histogram equalization, filtering and some morphological functions like erosion and dilation . Images used are RGB images and are converted into grayscale by eliminating the hue and saturation information while retaining the luminance. However, not all the image requires entire pre processing techniques. Some of them are good enough and no need to be improved using filtering or morphology. But, because of our research is to test the effectiveness of histogram equalization, this technique is compulsory conducted to all images. Figure 4 shows sample image of manuscript part after undergone through few preprocessing techniques.

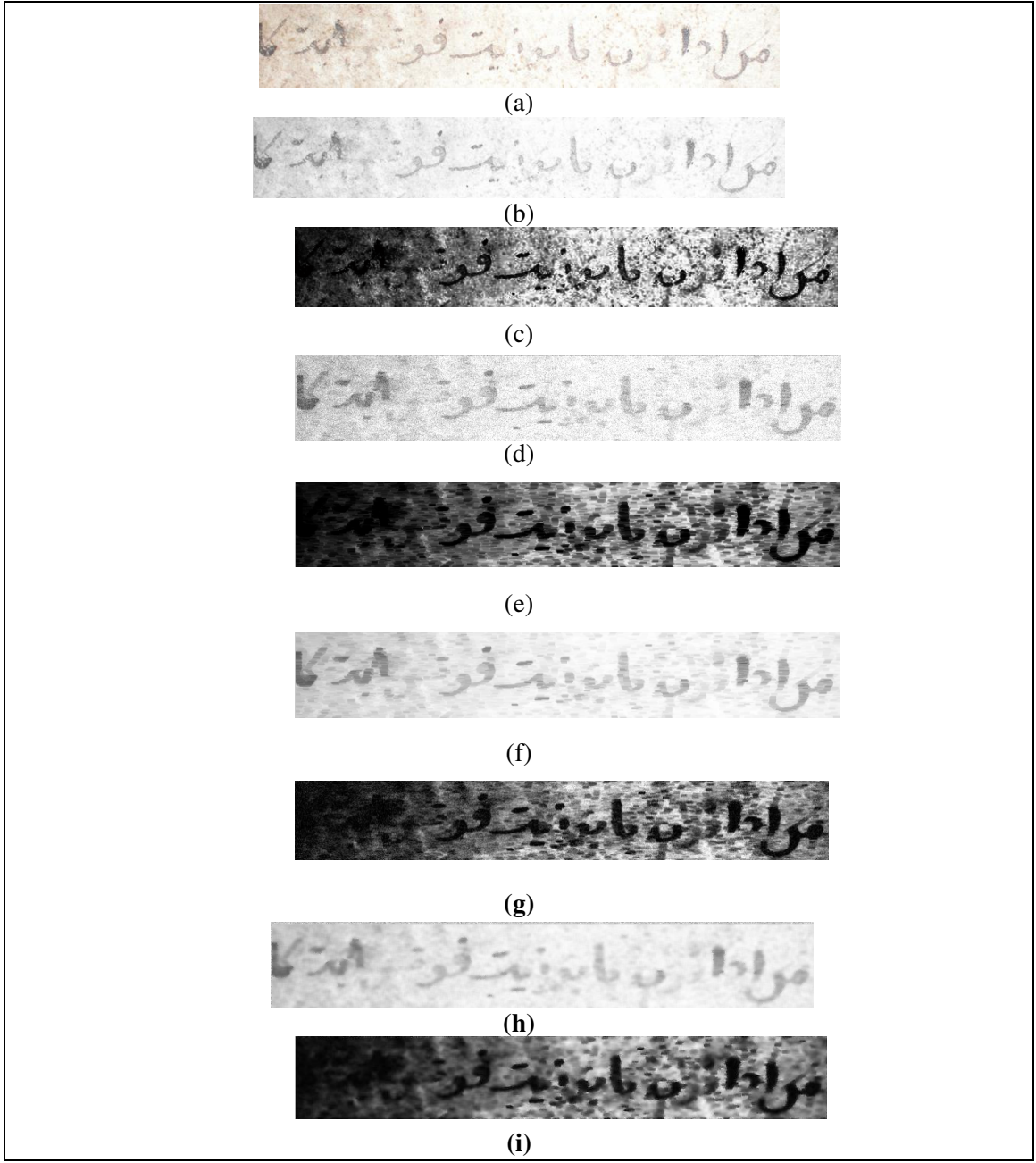

(a)

(b)

(c)

(d)

(e)

(f)

(g)

(h)

(i)

*Figure 4: (a) Original image. (b) Image converted to grayscale (c) Grayscale image after histogram equalization (d) Image 'b' after erosion (e) Image 'c' after erosion (f) Image 'd' after Gaussian Filter (g) Image 'e' after Gaussian Filter (h) Image 'd' after Wiener Filter (i) Image 'e' after Wiener Filter*

## 3. EXPERIMENT RESULT

Binarization

During this process, greyscale images were converted into binary images which means that every pixel in the image is convert to the binary values ("0" and "1")[11]. Here, we applied three techniques that probably representing the global and local threshold methods; that is Otsu's method[4], Niblack's[7] and Sauvola's Algorithm[9]. Figure 6 shows some of the result of these techniques to the old manuscript. These experiments were carried out on 16 samples of old manuscript with different cases from overall 36 images supplied by The Royal Museum of Pahang, Malaysia. Researcher was advised by evaluator to find the most distorted image among all. Besides, with human capability to visually inspect, it is impossible to conduct too many evaluations of data in short term period[12]. Furthermore, the historical document images consists Jawi characters may have complicated symbol that contribute to mis-interpretation.

By looking at the issues, we decided to separate the characters by two Units that is Link-symbol character and Single-symbol character. Figure 5 shows the example of this step.

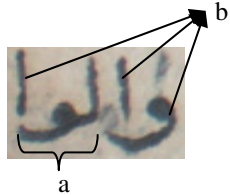

*Figure 5: (a) linked-symbol (b) single symbol*

A total of 1118 symbols were evaluated from these 16 images. Table 2 shows how the evaluation and assesment of each link-symbol and single symbol been assessed by the evaluator. Whichever symbol that 98%-100% could be recognized by evaluator, it will be marked as Recognized Symbol. The characteristics of Recognized symbol must meet the requirement of Jawi character that is; (1) clear dots in each symbol and (2) clear curve of each link-symbol or single-symbol. If one of these two conditions is unsure, the symbol will marked as unsure symbol. If none of the condition complied, the symbol is considered as Unknown Symbol.

From analysis of the results, the following conclusions can be reached:

1. Surprisingly, even our aim is to enhance the image quality by applying Histogram Equalization, each equalized image shows much more bad result than original grayscale image. From the findings, we can conclude that histogram equalization is not working in any historical document image type due to illumination and degraded paper problems.
2. The major problem of recognizing the symbol is to identify dots and curve in each symbol. It is because number of dots and curve form in each Jawi character represents different type of character and of course it will interpret different meaning.

3. Pre-processing technique that is erosion shows improvement towards the recognizing words of 12 from 16 images tested.

4. Filtering techniques such as Gaussian and Wiener filter also give impact to the quality of image. For comparison, Wiener Filter is better than Gaussian in terms of providing smooth and clear image [13][14].

5. For binarization techniques, by choosing the most suitable threshold and parameter values, results show that Niblack's and Sauvola's algorithm work well rather than global Otsu's method.

6. However, the computational cost indicated that Otsu's method utilized less time rather than other two local techniques. This is due to its less complicated algorithm with no extra parameters involved. For local approaches, with additional computation of the threshold using the dynamic range of image gray-value standard deviation, R showed that Sauvola's consumed much more time than Niblack's.

7. Global threshold binarization has the poorest result, performing worse than any of the local thresholding algorithms.

8. Niblack's algorithm with optimal parameters (93% overall achieved rate) performs better than the Sauvola-related algorithms (92% overall achieved rate) and Otsu's method (87% overall achieved rate).

| Technique | Image Symbol | 100% Recognized Symbol | Unsure Symbol | Unknown Symbol |
|---|---|---|---|---|
| Otsu's | 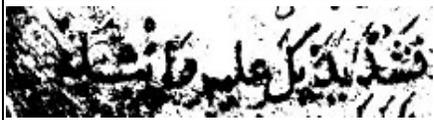 | فشد يذ يك علي | شد | 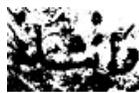 |
| Niblack's | 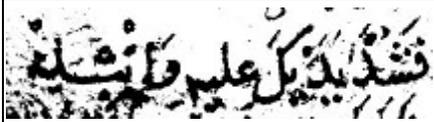 | فشد يذ يك علي | وانشد | - |
| Sauvola's | 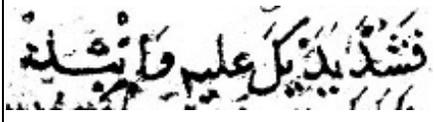 | فشد يذ يك علي | فانشدا | - |

*Table 1 : Example of Symbol Recognition by human assessment*

| Technique | Recognized + Unsure Symbol | | Rate (%) | | Overall Rate (%) | Average Computation Time (s) |
|---|---|---|---|---|---|---|
| | Linked-symbol | Single-symbol | Linked-symbol | Single-symbol | | |
| Otsu's | 741/857 | 232/261 | 86% | 89% | 87% | 3.1 |
| Niblack's | 803/857 | 240/261 | 94% | 92% | 93% | 3.9 |
| Sauvola's | 783/857 | 242/261 | 91% | 93% | 92% | 4.1 |

*Table 2 : Evaluation results (857 linked-symbol images and 261 single symbol images with total 1118 symbol images)*

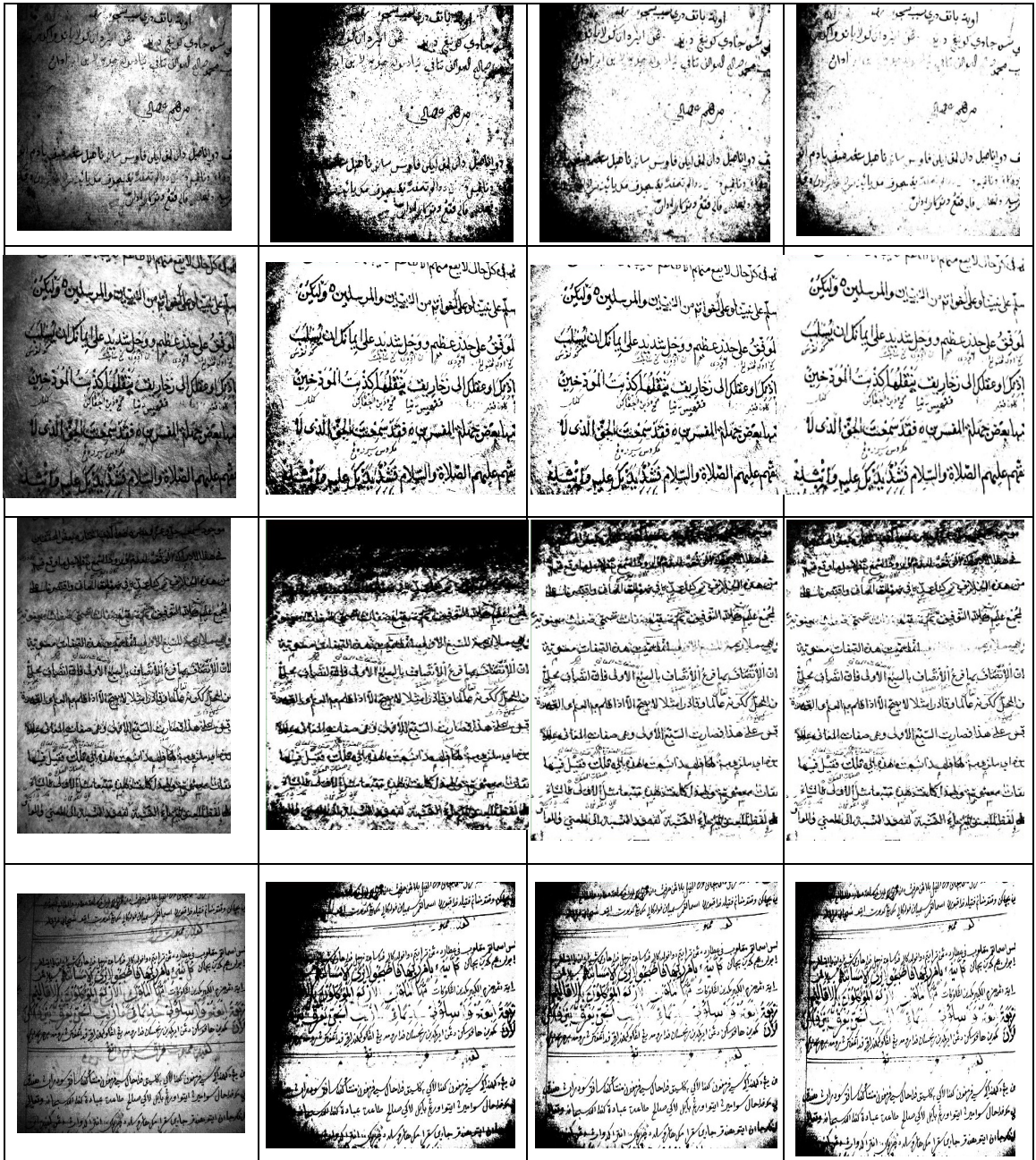

(a)  (b)  (c)  (d)

*Figure 6: (a) Image before binarization (b) Result of Otsu's method (c) Result of Niblack's Algorithm (d) Result of Sauvola's Algorithm*

## 4.  CONCLUSION

This paper has presented a comparison of several binarization algorithms by measuring their end-to-end word recognition performance on archive document word images. We have described algorithms that utilize spatial structure, global and local features or both. Many algorithms require extensive pre-processing steps in order to obtain useful data to work with because document image and data mining classification techniques is still in infancy. The conclusion is that no single algorithm works well for all types of image but some work well than others for particular types of images suggesting that improved performance can be obtained by automatic selection or combination of appropriate algorithm(s) for the type of document image under investigation. Also, researcher could improve the post processing step such as by adding edge detection techniques and further enhanced by an innovative image refinement technique and a formulation of a class proper method. Somehow, other alternative technique also could be tested like binarization using directional wavelet transforms hybrid thresholding and high performance adaptive binarization. Furthermore, to enhance the evaluation method, researcher should take into consideration for using Jawi Optical Character Recognition rather than human assessment due to human limits and view error.

# REFERENCES

[1]     Ergina Kavallieratou and   Stamatatos Stathis (2006). Adaptive Binarization of Historical Document Images. Proceedings of the 18th International Conference on Pattern Recognition (ICPR'06), pp 742-745

[2]     http://homepages.inf.ed.ac.uk/rbf/HIPR2/histeq.htm

[3]     Meng-Ling Feng and Yap-Peng Tan (2004). Adaptive Binarization Method For Document  Image Analysis. Proceedings of the 2004 IEEE International Conference on Multimedia and Expo(ICME)

[4]     N.Otsu (1979) A threshold selection method form gray-level histograms. Proceedings of the 1986 IEEE Transactions Systems, Man and Cybernetics, Vol.9, No. 1, 62-66

[5]     D.Trier and T. Taxt (1994). Evaluation of binarization methods for utility map images. Proceedings of the 1994 IEEE International Conference on Image Processing.

[6]     S. D. Yanowitz and A. M. Bruckstein (1989), A New Method for Image Segmentation,Computer Vision, Graphics and Image Processing

[7]     Niblack (1986), An Introduction to Digital Image Processing, pp. 115 - 116, Prentice Hall.

[8]     Zhang, Z. and C.L. Tan (2001). Restoration of images scanned from thick bound documents in Image Processing. Proceedings. 2001 International Conference.

[9]     J.Sauvola and M. Pietika Kinen (2000). Adaptive document image binarization. Pattern Recognition. Vol. 33 (2000), 225 – 236

[10]   http://www.digitalcamerareview.com/default.asp?newsID=3074&review=olympus+fe-23

[11]   Tushar Patnaik, Shalu Gupta, Deepak Arya(2010), Comparison of Binarization Algorithm in Indian  Language OCR, Proceedings of ASCNT – 2010, CDAC, Noida, India, pp. 61 – 69

[12]   Nigel Bivan (2009). The evaluation of accessibility, usability and user experience. The Universal Access Handbook, C Stepanidis (ed), CRC Press, 2009

[13]   H. B. Kekre and Dhirendra Mishra. (2010) Four Walsh Transform Sectors Feature Vectors for Image Retrieval from Image Databases, (IJCSIT) International Journal of Computer Science and Information Technologies, Vol. 1 (2) , pp. 33-37.

[14]   Y.Ramadevi, T.Sridevi, B.Poornima, B.Kalyanih. (2010) Segmentation And Object Recognition Using Edge Detection Techniques, International Journal of Computer Science & Information Technology (IJCSIT), Vol 2, No 6, December 2010, pp. 153-161.